\documentclass[11pt,a4paper]{article}
\usepackage{acl2021}
\usepackage{times}
\usepackage{latexsym}
\usepackage{multirow}
\usepackage{colortbl}
\usepackage{hhline}
\usepackage{graphicx}
\usepackage{stmaryrd}
\usepackage{subcaption}
\usepackage{diagbox}

\usepackage{xcolor}
\definecolor{best}{RGB}{1, 55, 0}
\definecolor{better}{RGB}{51, 105, 30}
\definecolor{good}{RGB}{51, 125, 30}
\definecolor{ok}{RGB}{124, 179, 66}
\definecolor{bad}{rgb}{0.8, 0.5, 0.}
\definecolor{worse}{rgb}{0.8, 0.0, 0.}

\newcommand{\bt}[1]{{\texttt {BT(#1)}}}
\newcommand{\ft}[1]{{\texttt {FT(#1)}}}

\newcommand{\btp}{{\texttt {BT}}}

\newcommand{\ftbt}[1]{{\ft{\bt{#1}}}}

\newcommand{\mono}[1]{\texttt {Mono$_{#1}$}}
\newcommand{\para}[1]{\texttt {Par$_{#1}$}}

\newcommand{\mb}[1]{} 
\newcommand{\af}[1]{} %
\newcommand{\kh}[1]{}

\usepackage{microtype}

\title{Fully Synthetic Data Improves Neural Machine Translation with Knowledge Distillation}
\author{Alham Fikri Aji \and
  Kenneth Heafield \\
  School of Informatics, University of Edinburgh \\
  10 Crichton Street \\
  Edinburgh EH8 9AB \\
  Scotland\\
  \texttt{a.fikri@ed.ac.uk, kheafiel@inf.ed.ac.uk}
}

\begin{document}
\maketitle
\begin{abstract}
This paper explores augmenting monolingual data for knowledge distillation in neural machine translation.
Source language monolingual text can be incorporated as a forward-translation. Interestingly, we find the best way to incorporate target language monolingual text is to translate it to the source language and round-trip translate it back to the target language, resulting in a fully synthetic corpus.
We find that combining monolingual data from both source and target languages yields better performance than a corpus twice as large only in one language.
Moreover, experiments reveal that the improvement depends upon the provenance of the test set.  
If the test set was originally in the source language (with the target side written by translators), then forward translating source monolingual data matters.  
If the test set was originally in the target language (with the source written by translators), then incorporating target monolingual data matters.  





\end{abstract}

\section{Introduction}

When we used knowledge distillation~\cite{ba2014deep, hinton2015distilling, kim2016sequence} to build faster neural machine translation (NMT)~\cite{bahdanau2014neural, NIPS2014_a14ac55a} models, the quality loss depended on the test set. Test sets collected in the source language then professionally translated to the target language showed little change in BLEU~\cite{papineni2002bleu}.  However, test sets collected in the target language then professionally translated to the source language showed larger losses in BLEU due to knowledge distillation. This led us to examine source and target monolingual data augmentation methods and their relationship with translationese.


Target monolingual data is normally incorporated by back translation~\cite{sennrich2016improving}, which adds a pseudo-parallel corpus of real target-language text paired with their machine translations to the source language. Sequence-level knowledge distillation uses forward translation: source language text is translated to the target language to form a pseudo-parallel corpus, which the student then learns from~\cite{kim2016sequence}.  The two methods can be composed: authentic text in the target language can be translated to the source language then translated again to the target language, resulting in a completely synthetic parallel corpus consisting solely of machine translation output.  Oddly, our experiments show that this completely synthetic corpus is better for training students in knowledge distillation compared to concatenating a back-translated corpus.  
We can incorporate source monolingual data by simply forward translating it with a teacher model.

This paper explores the interaction of monolingual data augmentation in knowledge distillation. We find that the BLEU improvement depends on the provenance of the test set. Augmenting target monolingual data improves BLEU for test set originated in target language. Inversely, source monolingual data improves BLEU for a test set originally in the source language. Augmenting both source and target monolingual data yields the best result.


We also investigate whether forward-translating seen data is necessary, as some research suggests forward-translating the same corpus used by the teacher~\cite{kim2016sequence, freitag2017ensemble, Yim_2017_CVPR}. We find that the student trained on new unseen monolingual data performs equally as the one with one trained on the same dataset as the teacher, as long as they share the same domain.

The amount of training data, including the augmented one affects model performance~\cite{edunov2018understanding, sennrich2019revisiting, araabi2020optimizing}. Therefore, we also explore the augmented monolingual data size. We find that augmenting more monolingual data is generally better. However, having more varied data based on the monolingual language origin is much more important.  Using monolingual data from both source and target languages is better than having more data in total from one language.  

\begin{figure}[h!]
\centering
\includegraphics[width=6cm]{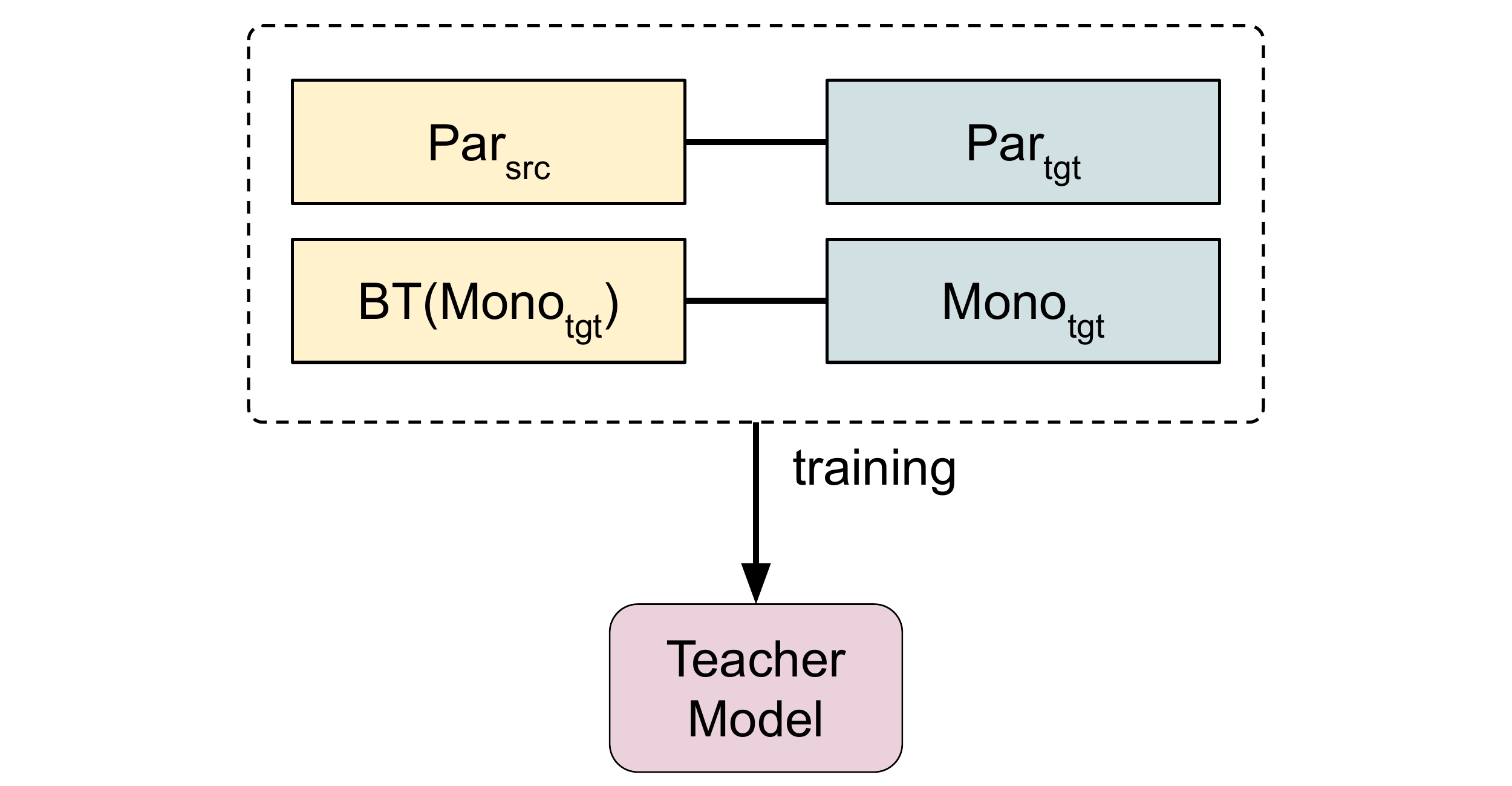}
\caption{Illustration for setting up a teacher model. The teacher is trained with parallel corpus + back-translation data. \label{illust-teacher}}

\includegraphics[width=6cm]{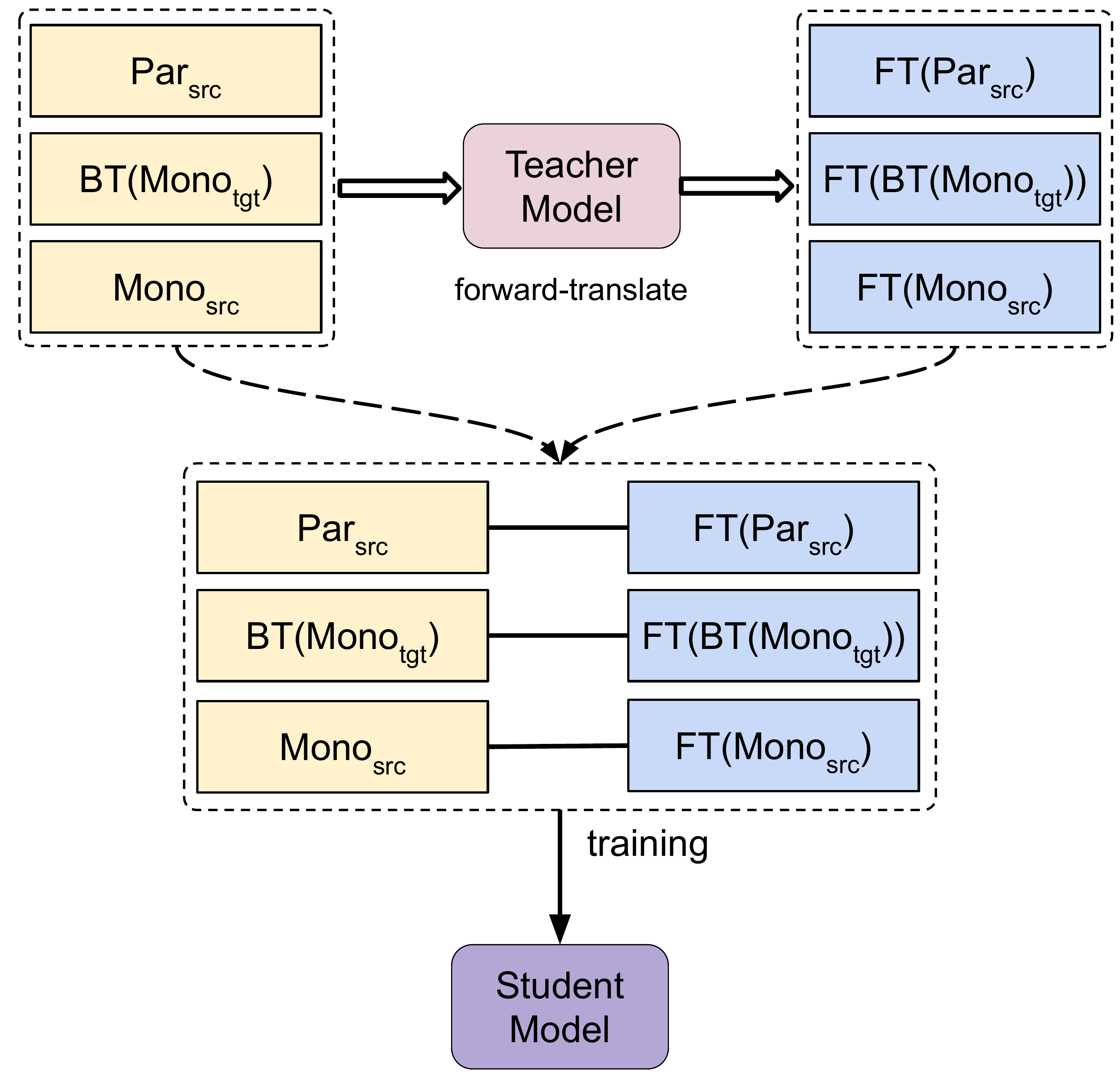}
\caption{Illustration for setting up a student model with interpolated knowledge distillation. The student is trained with forward-translated data by the teacher. \label{illust-student} 
}
\end{figure}

\section{Related Work}

\subsection{Knowledge Distillation}

Smaller neural models usually have an advantage in terms of efficiency~\cite{kim2019research, bogoychev2020edinburgh, aji2020compressing}. However, they are usually of poorer quality than large models. Smaller models have shown to perform better if trained with knowledge distillation~\cite{ba2014deep, hinton2015distilling, kim2016sequence} than trained from scratch, which we confirm. In knowledge distillation, the small model is trained by learning the output distribution of a larger model.

\citet{kim2016sequence} proposed interpolated knowledge distillation for sequence-to-sequence models. In interpolated knowledge distillation, students learn directly from the output produced by their teachers. This interpolated knowledge distillation is easy to apply, as practically we simply produce forward translated synthetic data using a teacher model once. Despite its simplicity, interpolated knowledge distillation has been shown to be useful for training small models without compromising quality~\cite{kim2019research, bogoychev2020edinburgh}. 

\subsection{Translationese Evaluation}

Human-written translations, sometimes called translationese, show different characteristics compared to naturally written text. The word distribution differs from natural text, as the translators are influenced by the original language when producing the translation~\cite{koppel2011translationese}. In the context of machine translation, sentences in the test set may originally come from the source language or the target language, which was then translated by annotators. 

Several studies have found that the performance of the NMT model is sensitive to translationese~\cite{zhang2019effect, bogoychev2019domain, edunov-etal-2020-evaluation}.
The ranking of a news translation shared task changes depending on which part of the test set is used~\cite{zhang2019effect}. The training data used also affects the performance of NMT. \citet{edunov-etal-2020-evaluation} have found that leveraging back translation data improves the performance of test sets derived from target-side data that are translated to source-side. \citet{bogoychev2019domain}~found otherwise that the forward translation synthesis data achieved better BLEU in the test set from the source-side. Given that our research in knowledge distillation makes use of monolingual data from both languages, we will conduct a performance evaluation based on the original source of the test set as well.

\section{Knowledge Distillation With Monolingual Data}

To perform a knowledge distillation, we first have to prepare a teacher model. The teacher model is trained as usual by using parallel data (\para{src} -- \para{tgt}) and back translation data (\bt{\mono{tgt}} -- \mono{tgt}). An illustration for setting up the teacher model can be seen in Figure~\ref{illust-teacher}.

In interpolated knowledge distillation, the student learns to mimic the teacher's translations.  So we take source language text, forward translate  (FT) it with the teacher, and train the student on a pseudo-parallel corpus of source text and the teacher's translations.  An illustration for this approach can be seen in Figure~\ref{illust-student}. There are several options for source-side monolingual data to be forward-translated by the teacher model, which we will explore further in this paper:
\begin{itemize}
    \item \textbf{Forward-translated source-side parallel data: \para{src} -- \ft{\para{src}}}  
    We can take the source side of the available parallel data. Naturally, this data is learnt by the teacher and follows prior work's~\cite{kim2016sequence, freitag2017ensemble, Yim_2017_CVPR} suggestion to use the same data for the student.
    
    \item \textbf{Forward-translated back translation: \bt{\mono{tgt}} -- \ftbt{\mono{tgt}}}  
    We can forward-translate back translation (BT) data (i.e., target-side monolingual data that has been translated to the opposite direction). Similar to the previous data, we can use the same BT data as the one used by the teacher. Though, we also explore using BT that teachers have never seen before. It should be noted that in this case, both the source and target are synthetic data.
    
    \item \textbf{Forward-translated source-side monolingual data: \mono{src} -- \ft{\mono{src}}} 
    Lastly, we can take advantage of the source-side monolingual data by directly translating it. Different from the previous data sources, this data has never been seen before by the teacher.

\end{itemize}

\begin{table}[ht!]
\small
    \centering
    \begin{tabular}{@{}lr@{ $\rightarrow$ }l@{}}
    \hline
        
         Training set & Source & Target \\ \hline
         \multicolumn{3}{c}{\textbf{Used by teacher}} \\
         \textbf{\para{}} & \para{src} & \para{tgt} \\
         \textbf{\btp} & \bt{\mono{tgt}} & \mono{tgt} \\
         \multicolumn{3}{c}{\textbf{Used by student}} \\
         \textbf{\ft{\para{}}} & \para{src} & \ft{\para{src}} \\
         \textbf{\ft{\btp}} & \bt{\mono{tgt}} & \ftbt{\mono{tgt}} \\
         \textbf{\ft{\mono{}}} & \mono{src} & \ft{\mono{src}} \\
         
    \hline
    \end{tabular}
    \caption{Dataset summary for the experiments.}
    \label{tab:data-summary}
\end{table}

To summarize, we define the following training sets, shown in Table~\ref{tab:data-summary}. Some research~\cite{kim2016sequence, freitag2017ensemble, Yim_2017_CVPR} suggests training the student model with the same data as the teacher model, namely~\textbf{\ft{\para{}}} and~\textbf{\ft{\btp}}. We also have access to source-side monolingual data which can be used as~\textbf{\ft{\mono{}}}. We will explore whether each of these datasets can be used in interpolated knowledge distillation.  Since prior work has shown that NMT evaluation is sensitive towards training data direction~\cite{bogoychev2019domain, edunov-etal-2020-evaluation}, we also split the test set based on the original language.

\section{Experiment Setup}

\subsection{Model Configuration}

Our teacher model uses a Transformer \texttt{Big} architecture~\cite{vaswani2017attention} with 6 encoder and decoder layers, with the embedding 1024 and a feed-forward network (FFN) size of 4096.  Following~\citet{bogoychev2020edinburgh}, most of our students use the \texttt{Tiny} architecture with 6 standard Transformer layers for the encoder and 2 RNN-based Simpler Simple Recurrent Unit (SSRU) layers~\cite{kim2019research} for the decoder. We also use a unit size of 256 and filter size of 1536. Beside the \texttt{Tiny} architecture, in section \ref{sec-size} we also explore student model of different layer sizes and unit sizes.

The text is pre-processed into subword units with SentencePiece~\cite{kudo2018sentencepiece}. We also use a tied-embeddings layer~\cite{press2017using}. The model is trained using the Marian toolkit~\cite{mariannmt} until no improvement is found for 20 consecutive validation steps. We evaluate once every 1k steps, and the training ends after ~200k steps. We use SacreBLEU~\cite{post2018call} to measure the model's performance.

Our model is trained on 4 GPUs\footnote{P100 or GeForce RTX 2080 Ti, depending on availability} with 10 GB dynamic mini-batch. Training time depends on the size of the model and data. Teacher model takes up to about 8 days, while student model can be trained in 1-2 days.

\subsection{Data}
\paragraph{Turkish-English (Tr-En)}
Tr-En parallel data comes from the WMT17 news translation task, which consists of 207k pairs. Monolingual English data for back-translation is collected from NewsCrawl 2017. Monolingual Turkish data for forward translation is collected from NewsCrawl 2018 and 2019.

\paragraph{Estonian-English (Et-En)}

Similarly, our Et-En parallel data comes from WMT18 news translation task (over-sampled to 10M pairs). English and Estonian monolingual data were obtained from NewsCrawl from 2017 to 2014. In addition, we used the BigEst Estonian Corpus for additional monolingual Estonian. We clean the data by removing sentences that are too short (less than 3 words) or too long (more than 150 words).

\paragraph{Spanish-English (Es-En)}

Lastly, the Es-En parallel data is a combination of data from ParaCrawl release-5 and Opus, totalling 102M pairs. Monolingual data for English is taken from NewsCrawl 2014-2017. Spanish monolingual data is a combination of NewsCrawl 2007-2018, Europarl v9, and Gigaword corpus.

\begin{table*}[ht!]

\centering

\begin{tabular}{@{}l@{ ~}l@{ }|@{ }c@{ ~ }c@{ ~ }c@{ }|c@{ ~ }c@{ ~ }c@{ }|c@{ ~ }c@{ ~ }c@{ }|ccc@{}}
\multicolumn{13}{l}{\textbf{BLEU by original language} }                                                                                                                                                                                                                                                                                       \\ 
\hline
& & \multicolumn{9}{c|}{Tr-En model}                                                                                                                                                                                                & \multicolumn{3}{c}{Et-En model}                                           \\
  &                             & \multicolumn{3}{@{ }c@{ }|}{WMT16}                                                & \multicolumn{3}{@{ }c|}{WMT17}                                                & \multicolumn{3}{c|}{WMT18}                                                & \multicolumn{3}{c}{WMT13}                                                 \\
  & Test original language=                              & src                      & tgt                     & all                   & src                      & tgt                      & all                   & src                      & tgt                      & all                   & src                      & tgt                      & all                   \\ 
\hline \hline
 \parbox[t]{2mm}{\multirow{8}{*}{\rotatebox[origin=c]{90}{\small Training Data}}} & \textbf{Big Model  (781 MB)}                                                                                           & \multicolumn{1}{l}{}    & \multicolumn{1}{l}{}    & \multicolumn{1}{l|}{} & \multicolumn{1}{l}{} & \multicolumn{1}{l}{}    & \multicolumn{1}{l|}{}    & \multicolumn{1}{l}{} & \multicolumn{1}{l}{} & \multicolumn{1}{l|}{}    & \multicolumn{1}{l}{}    & \multicolumn{1}{l}{} & \multicolumn{1}{l}{}  \\
& ~~\para{} + \btp                             & \color{best} 19.7                    & \color{best} 28.2                    & \color{best} 24.4                  & \color{best} 18.7                    & \color{best} 28.4                    & \color{best} 23.9                  & \color{best} 16.1                    & \color{best} 32.2                    & \color{best} 25.0                  & \color{best} 28.2                    & \color{best} 38.2                    &\color{best}  33.6                  \\
& \textbf{Tiny Model (65 MB)}                     & \multicolumn{1}{l}{}    & \multicolumn{1}{l}{}    & \multicolumn{1}{l|}{} & \multicolumn{1}{l}{}    & \multicolumn{1}{l}{}    & \multicolumn{1}{l|}{} & \multicolumn{1}{l}{}    & \multicolumn{1}{l}{}    & \multicolumn{1}{l|}{} & \multicolumn{1}{l}{}    & \multicolumn{1}{l}{}    & \multicolumn{1}{l}{}  \\
& ~~\para{} + \btp                              & \color{worse} 12.9                    & \color{worse} 17.0                    & \color{worse} 15.1                  & \color{worse} 12.3                    & \color{worse} 17.6                    & \color{worse} 15.0                  & \color{worse} 10.5                    & \color{worse} 19.2                    & \color{worse} 15.1                  &\color{worse}  25.1                    &\color{worse}  32.9                    &\color{worse}  29.5                  \\

& ~~\ft{\para{}+\btp}                              & \color{bad} {17.8}   & \color{ok} {24.2}  & \color{bad} 21.3                  & \color{bad} {16.6}   & \color{ok}{24.8}  & \color{bad} 20.7                  & \color{bad} {14.3}   & \color{ok}{27.8}  & \color{bad} 21.6                  & \color{bad} {26.4}   & \color{ok}{33.4}  & \color{bad} 30.3                  \\
& ~~\ft{\para{}+\mono{}}                              & \color{good}{19.2}  & \color{bad} {22.3}   & \color{bad} 21.1                  & \color{good}{18.3}  & \color{bad} {23.6}   & \color{good} \color{bad} 21.1                  & \color{best}{16.0}  & \color{bad} {26.6}   & \color{bad} 21.8                  & \color{best}{28.6}  & \color{bad} {31.7}   & \color{bad} 30.6                  \\
& ~~\ft{\para{}+\mono{}+\btp}           & \color{best}{19.7}  & \color{good}{25.0}  & \color{good} 22.6                  & \color{best}{18.9}  & \color{good}{25.8}  & \color{good} 22.5                  & \color{best}{16.2}  & \color{good}{29.0}  & \color{good} 23.1                  & \color{best}{28.1}  & \color{good}{34.5}  &  \color{good} 31.6                  \\
& ~~\ft{\para{}+\mono{}}\texttt{+}\btp & \color{good} 19.2  & \color{ok} 24.3 & \color{ok} 22.2 & \color{best} 18.6 & \color{ok} 24.9  & \color{ok} 22.0 & \color{good} 15.7 & \color{ok} 27.6 & \color{ok} 22.4 & \color{good} 27.8 & \color{ok} 33.1 & \color{ok} 30.9 \\
\hline
\end{tabular}

\rule{0pt}{2ex}

\begin{tabular}{@{}l@{ ~}l@{ }|@{ }c@{ ~ }c@{ ~ }c@{ ~ }c@{ }|c@{ ~ }c@{ ~ }c@{ ~ }c@{ }|c@{ ~ }c@{ ~ }c@{ ~ }c@{}} 
\hline
 &                                & \multicolumn{12}{c}{Es-En model}                                                                                                                                                                                                                                                                       \\
 &                             & \multicolumn{4}{c|}{WMT10}                                                                       & \multicolumn{4}{c|}{WMT11}                                                                       & \multicolumn{4}{c}{WMT12}                                                                        \\
& Test original language=& src                      & tgt                      & other                & all                   & src                      & tgt                      & other                & all                   & src                      & tgt                      & other                & all                   \\ 
\hline \hline
 \parbox[t]{2mm}{\multirow{8}{*}{\rotatebox[origin=c]{90}{\small Training Data}}} &\textbf{Big Model (781 MB)}                                                                                             & \multicolumn{1}{l}{}    & \multicolumn{1}{l}{}    & \multicolumn{1}{l}{} & \multicolumn{1}{l|}{} & \multicolumn{1}{l}{}    & \multicolumn{1}{l}{}    & \multicolumn{1}{l}{} & \multicolumn{1}{l|}{} & \multicolumn{1}{l}{}    & \multicolumn{1}{l}{}    & \multicolumn{1}{l}{} & \multicolumn{1}{l}{}  \\
&~~\para{} + \btp                                                                                                     & \color{best} 43.9                    & \color{best} 57.2                    & \color{best} 27.7                 & \color{best} 37.8                  & \color{best} 41.1                    & \color{best} 54.1                    & \color{best} 25.7                 & \color{best} 35.7                  & \color{best} 53.0                    & \color{best} 59.9                    & \color{best} 25.5                 & \color{best} 39.9                  \\
&\textbf{Tiny Model (65 MB)}                                                                                            & \multicolumn{1}{l}{}    & \multicolumn{1}{l}{}    & \multicolumn{1}{l}{} & \multicolumn{1}{l|}{} & \multicolumn{1}{l}{}    & \multicolumn{1}{l}{}    & \multicolumn{1}{l}{} & \multicolumn{1}{l|}{} & \multicolumn{1}{l}{}    & \multicolumn{1}{l}{}    & \multicolumn{1}{l}{} & \multicolumn{1}{l}{}  \\
&~~\para{} + \btp                                                                                                     & \color{worse} 38.7                    & \color{bad} 53.1                    & \color{worse} 25.1                 & \color{worse} 34.4                  & \color{worse} 36.2                    & \color{worse} 49.4                    & \color{worse}  23.4                 & \color{worse} 32.3                  & \color{worse} 47.2                    & \color{worse} 54.3                    & \color{worse} 23.8                 & \color{worse}  36.5                  \\

&~~\ft{\para{}+\btp} & \color{bad} {41.8}  & \color{ok} {54.4} & \color{ok} 26.0                      & \color{bad} 35.7                     & \color{bad} {38.9}  & \color{good} 51.4 &  \color{ok} 24.0 & \color{bad} 33.7                     & \color{bad}51.2  & \color{ok}{56.1} & \color{ok}24.6                      & \color{ok} 38.2                     \\
&~~\ft{\para{}+\mono{}}                                                                                                     & \color{good}{42.9}  & \color{worse}{52.1}   & \color{good}26.6                 & \color{good}36.0                  & \color{best}{41.2}  & \color{worse}{49.4}   & \color{good}24.5                 & \color{good}34.3                  & \color{best}{52.8}  & \color{bad}{54.7}   & \color{good}25.0                 & \color{good}38.5                  \\
&~~\ft{\para{}+\mono{}+\btp}                                                                                  & \color{good}{42.7}  & \color{ok}{54.4}  & \color{good} 26.4                 & \color{good} 36.1                  & \color{best}{41.4}  & \color{ok}{50.4}  & \color{good}24.4                 & \color{good}34.4                  & \color{good}{52.3}  & \color{ok}{56.0}  & \color{good}24.8                 & \color{good}38.6                  \\
& ~~\ft{\para{}+\mono{}}\texttt{+}\btp & \color{good}42.9 & \color{ok} 54.1 & \color{ok} 26.1 & \color{good} 36.0 & \color{best}41.0 & \color{ok} 50.0 & \color{ok} 24.1 & \color{ok} 34.1 & \color{ok} 51.8 & \color{bad} 54.7 & \color{good} 24.8 & \color{ok} 38.1 \\
\hline
\end{tabular}

\caption{Experiment results in terms of BLEU, divided based on the test-set's original language. The Big model is our teacher model. Models trained with forward-translated data (denoted with FT(...)) are our student models. For the training data, please refer to Table~\ref{tab:data-summary}.}
\label{main-res}
\end{table*}

\section{Experiments and Discussion}

\subsection{Monolingual Data for Knowledge Distillation}

We start the experiment by training several knowledge-distilled systems on different subsets of training data according to Table~\ref{tab:data-summary}.
As a baseline, we also compare the performance with our \texttt{Big} teacher model, and \texttt{Tiny} non-distilled model. Both models are trained with the same training set of parallel corpus and back-translation corpus. The results can be seen in Table~\ref{main-res}. Following ~\citet{bogoychev2019domain}, we break down the test set by the language in which the text was originally written.

As expected, the quality of the non-distilled \texttt{Tiny} model is much lower than the \texttt{Big} model. This result confirms that knowledge distillation significantly improves the quality of smaller model. 

The students vary in performance depending on the data set and the original language of the test set.  The \texttt{Tiny} student model trained with parallel data and back-translation (\ft{\para{}~+~\btp}) performs well on test sets originating from the target language, but poorly on test sets originating from  the source language. In contrast, the model trained with parallel data and source-side monolingual data (\ft{\para{}~+~\mono{}}) works well on test sets originating from the source language, but poorly if originating from the target language. Combining all sets (\ft{\para{}~+~\mono{}~+~\btp}) generally yields the best result, regardless of the data origin. Since these patterns are consistent among different test sets, hence we simply report the averaged BLEU for the next experiments.

Older Workshop on Machine Translation campaigns created test sets by gathering text in a variety of languages then translating to all the other languages pivoting through English.  So, for example, a sentence originally written in Czech was translated to English and again from English to Spanish, forming part of the Spanish--English test set.  We label these cases as ``other'' in Table~\ref{main-res}. The model trained with \ft{\para{}~+~\mono{}} achieves marginally better BLEU than our recommended \ft{\para{}~+~\mono{}~+~BT}. We are unsure why, but the difference is small enough relative to the size of the test set to conclude anything.

We try two ways to combine forward translation and back translation: composition and concatenation.  In composition, target language monolingual text is translated to the source language then translated again to the target language in a round-trip.  Curiously, this \ft{\btp} data is fully synthetic. In contrast, we can simply concatenate the back-translated and forward-translated parallel corpora. The result in Table~\ref{main-res} (Last row) shows that apparently using \ft{\btp} is slightly better than \btp. We further confirm that this improvement significant, according to~\cite{koehn2004statistical} (p-value 0.002). Hence, we recommend to use composition: fully synthetic forward-translated back-translation data.  This makes sense because concatenation bypasses the teacher and exposes the student to real target language data, which is more difficult to learn from.  


\subsection{Using Unseen Monolingual Data}

There is a confounding factor in the use of source monolingual data: it was never seen by the teacher and, consequently, did not have a chance to overfit.  Separating this confound directly would require inventing a way to use source data to train the teacher. But we can test a related question: does it matter if the teacher and student use the same target monolingual data? 

To recall, the teacher model is trained with the parallel corpus (\para{}) and back translation data (\btp).  Following results from the previous section, to train the student we back translate the target monolingual data and then forward translate it to form a synthetic parallel corpus (\ft{\btp}). Both teacher and student require \btp~~data, which is coming from the target monolingual data . Do we need to use the same target monolingual data for both of them?

As shown in Table~\ref{tab-data-bt}, we explore training students with different target monolingual data. We divide NewsCrawl 2017 into equally-sized chunks. The teacher is trained with back translation of the first chunk. One student is trained with forward translation of back translation from the same chunk.  Another student is trained on forward translated back translations from a second chunk of NewsCrawl 2017 (unseen by the teacher). We find that both students perform equally well.

Furthermore, we also train students using forward-translated back translation constructed from different corpora: NewsCrawl 2010, CommonCrawl, and OpenSubtitle. We find that among those, NewsCrawl 2010 works best (and performs comparably with NewsCrawl 2017), whereas OpenSubtitle achieves worst BLEU. From these results, we conclude that the training data for students does not have to be the same as the teacher, as long as the domain agrees. Though, using CommonCrawl or  Opensub\footnote{http://opus.nlpl.eu/download.php?f=OpenSubtitles/v2018/ mono/OpenSubtitles.raw.\{en,et\}.gz} monolingual data is still better than not using any monolingual data at all.

Likewise, we also explored the effects of selecting the forward-translated monolingual data (\ft{\mono{}}). However, different from the back translation that comes from the target-side monolingual data, the source-side monolingual data is not used when training the teacher. So, in this experiment we only explore the domain from the data. In Table~\ref{tab-data-mono}, we find that the best monolingual source-side data for student is the one with the same domain as the teacher (news domain) model.


\begin{table}
\centering

\begin{tabular}{@{}l@{ }ccc@{}}
\hline
{\bt{\mono{en}} source} & \multicolumn{3}{c}{\small {Tr$\shortrightarrow$En avg BLEU}}    \\
 & src & tgt    & all    \\ \hline \hline
None & \phantom{0}4.5 & \phantom{0}5.0 & \phantom{0}4.9 \\
 NewsCrawl-17\small(same as Teacher) & 16.2 & 25.6 & 21.2 \\
 NewsCrawl-17\small(diff. w/ Teacher) & 16.3 & 25.9 & 21.4 \\
 NewsCrawl-10 & 15.9 & 25.7 & 21.1 \\
 English CommonCrawl & 15.7 & 23.3 & 19.7 \\
Opensub v2018 & 10.6 & 14.8 & 12.8 \\ \hline

\end{tabular}
\caption{Exploring different back-translation data sources when the student model is trained with \ft{\para{}~+~\btp} training data. None is trained only with \ft{\para{}}.\label{tab-data-mono}}

\rule{0pt}{4ex}

\begin{tabular}{@{}lccc@{}}
\hline
{\mono{tr} source} & \multicolumn{3}{c}{{Tr$\rightarrow$En avg. BLEU}}    \\
 & {src} & {tgt}    & {all}    \\ \hline \hline
None & \phantom{0}4.5 & \phantom{0}5.0 & \phantom{0}4.9 \\
 NewsCrawl-18 + 19  & 17.8 & 24.1 & 21.3 \\
 Turkish CommonCrawl & 17.0 & 22.8 & 20.2 \\
Opensub v2018 & 11.4 & 15.8 & 13.3 \\ \hline

\end{tabular}
\caption{Exploring different source-side monolingual data when the student model is trained with \ft{\para{}~+~\mono{}} training data. None is trained only with \ft{\para{}}. \label{tab-data-bt}}

\end{table}

\subsection{Monolingual Data Size}
\label{sec-size}

\begin{table}[ht!]
\begin{subtable}{0.5\textwidth}

\centering
\begin{tabular}{rr|ccc}
\hline
   \multicolumn{2}{c|}{augment size}    & \multicolumn{3}{c}{{Tr$\shortrightarrow$En avg BLEU}}    \\ 
 \small \ft{\btp{}} & \small \ft{\mono{}} & src & tgt   & all    \\ \hline \hline
    5M    & 0    & 16.2    & \textbf{25.6} & 21.2    \\
    0    & 5M    & \textbf{17.9}    & 24.1 & 21.3    \\
    2.5M    & 2.5M    & \textbf{17.9}    & \textbf{25.8} & \textbf{22.0}    \\ \hline
    10M    & 0    & 16.5    & \textbf{26.4} & 21.8    \\
    0    & 10M    & \textbf{18.1}    & 25.2 & 21.9    \\
    5M    & 5M    & \textbf{18.3}    & \textbf{26.6} & \textbf{22.7}    \\ \hline
    10M    & 10M    & \textbf{18.2}    & \textbf{27.5} & \textbf{23.2}  \\  \hline

\end{tabular}
\caption{Tr$\shortrightarrow$En model.}
\end{subtable}

\rule{0pt}{1ex}

\begin{subtable}{0.5\textwidth}
\centering
\begin{tabular}{rr|ccc} 
\hline
       \multicolumn{2}{c|}{augment size }    & \multicolumn{3}{c}{{Et$\shortrightarrow$En avg BLEU}}    \\ 
 \small \ft{\btp{}} & \small \ft{\mono{}} & src & tgt   & all    \\ \hline \hline
    32M & 0         & 26.4 & \textbf{33.4} & 30.3 \\
    0    & 32M      & \textbf{28.6} & 31.7 & 30.6 \\
    16M    & 16M    & \textbf{28.1} & \textbf{33.6}  & \textbf{31.3} \\ \hline
    32M    & 32M    & \textbf{28.1} & \textbf{34.5} & \textbf{31.6} \\  \hline  
\end{tabular}
\caption{Et$\shortrightarrow$En model.}
\end{subtable}

\rule{0pt}{1ex}

\begin{subtable}{0.5\textwidth}
\centering
\begin{tabular}{rr|ccc} 
\hline
       \multicolumn{2}{c|}{augment size }    & \multicolumn{3}{c}{{Et$\shortrightarrow$En avg BLEU}}    \\ 
 \small \ft{\btp{}} & \small \ft{\mono{}} & src & tgt   & all    \\ \hline \hline
 
    96M & 0         & 44.1 & \textbf{53.9} & 35.9 \\
    0      & 86M    & \textbf{45.8} & 52.1 & \textbf{36.3}  \\
    43M     & 43M   & \textbf{45.5} & 52.8 & \textbf{36.2} \\ \hline
    96M    & 86M    & \textbf{45.6} & \textbf{53.6} & \textbf{36.4} \\ \hline 
\end{tabular}
\caption{Es$\shortrightarrow$En model.}
\end{subtable}
\caption{Evaluating model's performance over various monolingual dataset size. Models trained with mixture \ft{\btp{}} + \ft{\mono{}} generally perform best.}
\label{tab-size}
\end{table}

In Table~\ref{main-res}, the best student model is trained with the combination of forward-translated back-translation and forward-translated source-side monolingual data (\ft{\para{}~+~\mono{} +~\btp}). However, that model is trained with the most amount of data compared to others. In this subsection, we further explore the effect of back-translation and source-side monolingual data under more controlled data size, to remove any artefact of different data size from interfering with the result.

Generally, more training data often leads to better performance. However, in our case, mixing our synthetic data is more important, than
As shown in Table~\ref{tab-size}, given the same total amount of augmented data size, mixing back-translation and source-side monolingual data achieves the best BLEU, compared to the models trained exclusively on either data. In fact, our Tr-En model trained with 2.5M \ft{\btp{}} + 2.5M \ft{\mono{}} achieves overall BLEU comparable to the models trained with 10M data exclusively on either side (22.0 BLEU vs 21.9 BLEU). \citet{soto2020selecting} has shown similar result with back-translation data, where augmenting from multiple sources is better than relying on one source of data.


\begin{table*}[ht!]
    \centering
    \begin{tabular}{@{}lrrrccr@{~~}lr@{~~}lr@{~~}l@{}}
        \hline
        Model &  \multicolumn{1}{c}{Size} & Embed. & FFN & Enc./Dec. & Prec & \multicolumn{6}{c}{Tr$\rightarrow$En avg. BLEU} \\
        & (MB) & size & size & depth & (bits) & src & ($\Delta$) & tgt & ($\Delta$) & all & ($\Delta$) \\
        \hline \hline
        Big (Teacher) & 781 & 1024 & 4096 & 6/6 & 32 & 18.2	& &	29.6	& &	24.4 \\
  
        Base (non-distill) & 232 & 512 & 2048 & 6/6 & 32 & 18.1 & \small (-0.1) & 29.1 & \small (-0.5) & 24.0 & \small (-0.4) \\ 
        Base & 232 & 512 & 2048 & 6/6 & 32 & 18.3 & \small (\phantom{-}0.1) & 28.2 &  \small (-1.4) & 23.8 & \small (-0.6) \\  \hline
        Small & 83 & 256 & 1536 & 6/6 & 32 & 18.0 &	\small (-0.1) & 27.2	& \small (-2.4) & 23.0 & \small (-1.4) \\ 
        Tiny & 65 & 256 & 1536 & 6/2 & 32 & 18.3 & \small (\phantom{-}0.1) & 26.6 & \small (-3.0) & 22.7	& \small (-1.7) \\
        Micro & 47  & 256 & 1024 & 4/2 & 32 & 17.3 & \small (-0.9) & 24.9 & \small (-4.7) & 21.4 & \small (-3.0) \\
        \hline
        Small-8bit & 21 & 256 & 1536 & 6/6 & 8 & 18.0 & \small (-0.2) & 27.1 & \small (-2.5) & 22.9 & \small (-1.5) \\ 
        Tiny-8bit & 17 & 256 & 1536 & 6/2 & 8 & 
17.8 & \small (-0.4) & 25.9 & \small (-3.7) & 22.2 & \small (-2.2) \\
        Micro-8bit & 12  & 256 & 1024 & 4/2 & 8 & 
17.7 & \small (-0.5) & 24.1 & \small (-5.5) & 21.4 & \small (-3.0) \\
        \hline
        Small-4bit & 10 & 256 & 1536 & 6/6 & 4 & 17.7 & \small (-0.5) & 25.8 & \small (-3.8) & 22.3 & \small (-2.1) \\
        Tiny-4bit & 8 & 256 & 1536 & 6/2 & 4 & 
17.9 & \small (-0.3) & 25.2 & \small (-4.4) & 22.0 & \small (-2.4) \\
        Micro-4bit & 6  & 256 & 1024 & 4/2 & 4 & 
16.6 & \small (-1.6) & 22.8 & \small (-6.8) & 20.1 & \small (-4.3) \\
        \hline
    \end{tabular}
    \caption{Performances of different Tr$\rightarrow$En student sizes, according to test set's original language. The teacher and non-distilled models are trained with \para{}+\btp, whereas all student models are trained with ~~\ft{\para{}+\btp+\mono{}} data.}
    \label{model-size}
\end{table*}

\begin{table}[ht!]
\begin{subtable}{0.5\textwidth}
    \centering
    \begin{tabular}{@{}lr@{~}lr@{~}lr@{~}l@{}}
    \hline
        Model & \multicolumn{6}{c}{Et$\rightarrow$En avg. BLEU} \\
        & src & ($\Delta$) & tgt & ($\Delta$) & all & ($\Delta$) \\
        \hline \hline
        Big (Teacher) & 28.3 & & 38.2	& &	33.6	\\
        \hline
        Tiny & 28.1	& \small (-0.2) &	34.5	& \small (-3.7) & 	31.6 &	\small (-2.0) \\
        Micro & 26.3 & \small (-2.0) &	30.6 & \small (-7.6) &	28.6 & \small (-5.0) \\
        \hline
        Tiny-8bit & 28.5 & \small (0.2) & 33.5 & \small (-4.7) &  	31.3 & \small(-2.3) \\
        Micro-8bit & 26.6	& \small (-1.7) & 	30.3 & \small(-7.9) & 	28.7 & \small(-4.9)\\
        \hline
    \end{tabular}
    \caption{Et$\shortrightarrow$En model.}
    \label{model-size-eten}
\end{subtable}    
\rule{0pt}{1ex}

\begin{subtable}{0.5\textwidth}
    \centering
    \begin{tabular}{@{}lr@{~}lr@{~}lr@{~}l@{}}
    \hline
        Model & \multicolumn{6}{c}{Es$\rightarrow$En avg. BLEU} \\
        & src & ($\Delta$) & tgt & ($\Delta$) & all & ($\Delta$) \\
        \hline \hline
        Big (Teacher) & 46.0 & & 57.1 & & 37.8\\
        \hline
        Tiny & 45.5 & \small (-0.5) & 53.6 & \small(-3.5) & 36.4 & \small(-1.4)\\
        Micro& 44.7 & \small (-1.3) & 51.3 & \small (-5.8) & 35.3 & \small(-2.5) \\
        \hline
        Tiny-8bit  & 45.5 & \small (-0.5) & 53.4 & \small (-3.7) & 	36.3 &	\small (-1.5) \\
        Micro-8bit & 44.6 & \small (-1.4) & 51.2 & \small (-5.9) & 35.1 &	\small (-2.7) \\
        \hline
    \end{tabular}
    \caption{Es$\shortrightarrow$En model.}
    \label{model-size-esen}
\end{subtable}
\caption{Performances of different student sizes, for Et$\rightarrow$En and Es$\rightarrow$En models according to test set's original language. Similarly, student models are trained with ~~\ft{\para{}+\btp+\mono{}} data.}
\label{model-size-etes}
\end{table}

\subsection{Exploring Different Student Sizes}

From Table~\ref{main-res}, we only see small BLEU-gap between teachers and students in the test set originating from the source. In some cases, we find that students perform equally or even better than teachers in that test set.
On contrary, students lose BLEU significantly (can be more than $-5$ points) when handling a test set originally written in the target language.  In this subsection, we confirm this pattern by training student models with different sizes.

We explore 4 variations of student size (\texttt{Base, Small, Tiny, and Micro}) as shown in Table~\ref{model-size}. The \texttt{Base} model uses Transformer encoder and decoder, whereas the other models use RNN-based Simpler Simple Recurrent Unit (SSRU) for decoders. We also stack the students with 8-bit fixed point quantization~\cite{kim2019research} and 4-bit log quantization~\cite{aji2020compressing} to achieve even smaller model size. Our 8-bit and 4-bit models are trained with full 32-bit precision first before continuing training under lower precision. 

The result shown in Table~\ref{model-size} confirms that the student models handle test sets with original source sentences better, compared to the opposite side. A Similar pattern can be observed for both Et$\shortrightarrow$En and En$\shortrightarrow$En models, as shown in Table~\ref{model-size-etes}. In our extremely small 4-bit \texttt{Micro} student, we ``only'' lose $-1.6$ BLEU points on the test set originating in the source language compared to $-6.8$ BLEU points on the test set originated in the target language. Moreover, our largest student (\texttt{Base}) already loses $-1.4$ BLEU points on that target-originated test set.

Model quantization can shrink the size more effectively than just using a smaller network. For example, the \texttt{Micro} model is bigger (47MB, 21.4 BLEU), but not as good as the \texttt{Small}-4bit model (10MB, 22.3 BLEU). Curiously, it seems that the model quantization also had a problem dealing with the test set originating from the target language. Meanwhile, the model quantization did not appear to have an impact on the test set performance from the source language, except for the \texttt{Tiny}-4bit which we suspected was due to its extremely small size.

We train a non-distilled \texttt{Base} model, to find out whether this performance decline is due to knowledge distillation or the small size of the model. As in Table~\ref{model-size}, we see that the distilled \texttt{Base} model yields worse BLEU on target-originated test ($-1.4$ BLEU) compared to the non-distilled variant ($-0.5$ BLEU). Therefore we confirm that distilled models somehow have difficulty handling a test set originally written in the target language.



\subsection{Student Output vs Translationse}


Student models perform better when the source text is original and the target text is translationese.  This makes some intuitive sense: all student training data is forward translated from some source sentence, just as translationese is forward translations.  To make this comparison more rigorous, we examined indica of translationese from the literature.


Translationese differs from original text in type-token ratio (TTR), length, and part-of-speech tag rate~\cite{lembersky2012adapting, daems2017translationese}. In Table~\ref{tab:ttr}, we show that indeed the student models are closer to translationese in type-token ratio than their teacher.  However, we found no difference in text length and part-of-speech tag rate.  


\begin{table}[]
    \centering
    \begin{tabular}{lcc}
    \hline
       & \multicolumn{2}{c}{origlang=} \\
        Generated by & tr & en \\ 
            \hline
            \hline
        Human reference & 0.151 & 0.176 \\
        Big model (teacher) & 0.164 & 0.164\\
        Base model (non-distil) & 0.164 & 0.165 \\
        Base model (student) & 0.158 & 0.163 \\
        Tiny model (student) & 0.158 & 0.160 \\
        \hline
    \end{tabular}
    \caption{Type-token ratio of English translation produced by our model and human reference, divided based on the input's original language. For human reference, Translation originated from English is the natural text, whereas translation originated from Turkish is the translationese.}
    \label{tab:ttr}
\end{table}

Based on the human reference from Table~\ref{tab:ttr}, we can see that human translationese shows lower TTR, compared to the natural text. Interestingly, our non-distilled \texttt{Big} and \texttt{Base} model show the same TTR regardless the original language. Performing knowledge distillation reduces TTR, confirming that the student models are more biased. This result is also consistent with~\cite{vanmassenhove2019lost, toral2019post}, where translation output is losing its linguistic richness. Deeper investigation towards model's output would be interesting. For example, by using human evaluation to analyze model output directly.

\section{Conclusion}

We have conducted experiments on data augmentation in NMT with knowledge distillation through: (1) Forward translating source-originated text, and (2) Forward translating back-translated target-originated text.
We found that both types of augmentation data had an impact on performance (in terms of BLEU score), depending on the test set's original language. Forward translating source-originated text worked well if the test set was also originated from the source language. In contrast, forward translating back translation data worked well if the test set was originated from the target language. Combining both data achieved the best overall performance, even under the same total data size. For example, a student trained with 5M of data (1) + 5M of data (1) achieved overall better BLEU compared to a student trained with 10M of only with data (1) or (2) exclusively.

Prior work often used the same back translation data for the teacher and the student. However, we found that this is not required, provided that the domain is the same. 
Since our student is trained on forward-translated data, our student models are more robust on handling the test set originated from the source language. Such test set is essentially a forward translation as well. Our 8 MB student model degraded only -0.5 BLEU compared to its 781 MB teacher model.




\bibliography{acl2020}

\begin{thebibliography}{30}
\expandafter\ifx\csname natexlab\endcsname\relax\def\natexlab#1{#1}\fi

\bibitem[{Aji and Heafield(2020)}]{aji2020compressing}
Alham~Fikri Aji and Kenneth Heafield. 2020.
\newblock Compressing neural machine translation models with 4-bit precision.
\newblock In \emph{Proceedings of the Fourth Workshop on Neural Generation and
  Translation}, pages 35--42.

\bibitem[{Araabi and Monz(2020)}]{araabi2020optimizing}
Ali Araabi and Christof Monz. 2020.
\newblock Optimizing transformer for low-resource neural machine translation.
\newblock In \emph{Proceedings of the 28th International Conference on
  Computational Linguistics}, pages 3429--3435.

\bibitem[{Ba and Caruana(2014)}]{ba2014deep}
Jimmy Ba and Rich Caruana. 2014.
\newblock Do deep nets really need to be deep?
\newblock \emph{Advances in neural information processing systems},
  27:2654--2662.

\bibitem[{Bahdanau et~al.(2014)Bahdanau, Cho, and Bengio}]{bahdanau2014neural}
Dzmitry Bahdanau, Kyunghyun Cho, and Yoshua Bengio. 2014.
\newblock Neural machine translation by jointly learning to align and
  translate.
\newblock \emph{arXiv preprint arXiv:1409.0473}.

\bibitem[{Bogoychev et~al.(2020)Bogoychev, Grundkiewicz, Aji, Behnke, Heafield,
  Kashyap, Farsarakis, and Chudyk}]{bogoychev2020edinburgh}
Nikolay Bogoychev, Roman Grundkiewicz, Alham~Fikri Aji, Maximiliana Behnke,
  Kenneth Heafield, Sidharth Kashyap, Emmanouil-Ioannis Farsarakis, and Mateusz
  Chudyk. 2020.
\newblock Edinburgh’s submissions to the 2020 machine translation efficiency
  task.
\newblock In \emph{Proceedings of the Fourth Workshop on Neural Generation and
  Translation}, pages 218--224.

\bibitem[{Bogoychev and Sennrich(2019)}]{bogoychev2019domain}
Nikolay Bogoychev and Rico Sennrich. 2019.
\newblock Domain, translationese and noise in synthetic data for neural machine
  translation.
\newblock \emph{arXiv preprint arXiv:1911.03362}.

\bibitem[{Daems et~al.(2017)Daems, De~Clercq, and
  Macken}]{daems2017translationese}
Joke Daems, Orph{\'e}e De~Clercq, and Lieve Macken. 2017.
\newblock Translationese and post-editese: How comparable is comparable
  quality?
\newblock \emph{Linguistica Antverpiensia New Series-Themes in Translation
  Studies}, 16:89--103.

\bibitem[{Edunov et~al.(2018)Edunov, Ott, Auli, and
  Grangier}]{edunov2018understanding}
Sergey Edunov, Myle Ott, Michael Auli, and David Grangier. 2018.
\newblock Understanding back-translation at scale.
\newblock In \emph{Proceedings of the 2018 Conference on Empirical Methods in
  Natural Language Processing}, pages 489--500.

\bibitem[{Edunov et~al.(2020)Edunov, Ott, Ranzato, and
  Auli}]{edunov-etal-2020-evaluation}
Sergey Edunov, Myle Ott, Marc{'}Aurelio Ranzato, and Michael Auli. 2020.
\newblock \href {https://doi.org/10.18653/v1/2020.acl-main.253} {On the
  evaluation of machine translation systems trained with back-translation}.
\newblock In \emph{Proceedings of the 58th Annual Meeting of the Association
  for Computational Linguistics}, pages 2836--2846, Online. Association for
  Computational Linguistics.

\bibitem[{Freitag et~al.(2017)Freitag, Al-Onaizan, and
  Sankaran}]{freitag2017ensemble}
Markus Freitag, Yaser Al-Onaizan, and Baskaran Sankaran. 2017.
\newblock Ensemble distillation for neural machine translation.
\newblock \emph{arXiv preprint arXiv:1702.01802}.

\bibitem[{Hinton et~al.(2015)Hinton, Vinyals, and Dean}]{hinton2015distilling}
Geoffrey Hinton, Oriol Vinyals, and Jeff Dean. 2015.
\newblock Distilling the knowledge in a neural network.
\newblock \emph{arXiv preprint arXiv:1503.02531}.

\bibitem[{Junczys-Dowmunt et~al.(2018)Junczys-Dowmunt, Grundkiewicz, Dwojak,
  Hoang, Heafield, Neckermann, Seide, Germann, Fikri~Aji, Bogoychev, Martins,
  and Birch}]{mariannmt}
Marcin Junczys-Dowmunt, Roman Grundkiewicz, Tomasz Dwojak, Hieu Hoang, Kenneth
  Heafield, Tom Neckermann, Frank Seide, Ulrich Germann, Alham Fikri~Aji,
  Nikolay Bogoychev, Andr\'{e} F.~T. Martins, and Alexandra Birch. 2018.
\newblock \href {http://www.aclweb.org/anthology/P18-4020} {Marian: Fast neural
  machine translation in {C++}}.
\newblock In \emph{Proceedings of ACL 2018, System Demonstrations}, pages
  116--121, Melbourne, Australia. Association for Computational Linguistics.

\bibitem[{Kim and Rush(2016)}]{kim2016sequence}
Yoon Kim and Alexander~M Rush. 2016.
\newblock Sequence-level knowledge distillation.
\newblock In \emph{Proceedings of the 2016 Conference on Empirical Methods in
  Natural Language Processing}, pages 1317--1327.

\bibitem[{Kim et~al.(2019)Kim, Junczys-Dowmunt, Hassan, Aji, Heafield,
  Grundkiewicz, and Bogoychev}]{kim2019research}
Young~Jin Kim, Marcin Junczys-Dowmunt, Hany Hassan, Alham~Fikri Aji, Kenneth
  Heafield, Roman Grundkiewicz, and Nikolay Bogoychev. 2019.
\newblock From research to production and back: Ludicrously fast neural machine
  translation.
\newblock \emph{EMNLP-IJCNLP 2019}, page 280.

\bibitem[{Koehn(2004)}]{koehn2004statistical}
Philipp Koehn. 2004.
\newblock Statistical significance tests for machine translation evaluation.
\newblock In \emph{Proceedings of the 2004 conference on empirical methods in
  natural language processing}, pages 388--395.

\bibitem[{Koppel and Ordan(2011)}]{koppel2011translationese}
Moshe Koppel and Noam Ordan. 2011.
\newblock Translationese and its dialects.
\newblock In \emph{Proceedings of the 49th Annual Meeting of the Association
  for Computational Linguistics: Human Language Technologies}, pages
  1318--1326.

\bibitem[{Kudo and Richardson(2018)}]{kudo2018sentencepiece}
Taku Kudo and John Richardson. 2018.
\newblock Sentencepiece: A simple and language independent subword tokenizer
  and detokenizer for neural text processing.
\newblock In \emph{Proceedings of the 2018 Conference on Empirical Methods in
  Natural Language Processing: System Demonstrations}, pages 66--71.

\bibitem[{Lembersky et~al.(2012)Lembersky, Ordan, and
  Wintner}]{lembersky2012adapting}
Gennadi Lembersky, Noam Ordan, and Shuly Wintner. 2012.
\newblock Adapting translation models to translationese improves smt.
\newblock In \emph{Proceedings of the 13th Conference of the European Chapter
  of the Association for Computational Linguistics}, pages 255--265.

\bibitem[{Papineni et~al.(2002)Papineni, Roukos, Ward, and
  Zhu}]{papineni2002bleu}
Kishore Papineni, Salim Roukos, Todd Ward, and Wei-Jing Zhu. 2002.
\newblock Bleu: a method for automatic evaluation of machine translation.
\newblock In \emph{Proceedings of the 40th annual meeting of the Association
  for Computational Linguistics}, pages 311--318.

\bibitem[{Post(2018)}]{post2018call}
Matt Post. 2018.
\newblock A call for clarity in reporting bleu scores.
\newblock \emph{arXiv preprint arXiv:1804.08771}.

\bibitem[{Press and Wolf(2017)}]{press2017using}
Ofir Press and Lior Wolf. 2017.
\newblock Using the output embedding to improve language models.
\newblock In \emph{Proceedings of the 15th Conference of the European Chapter
  of the Association for Computational Linguistics: Volume 2, Short Papers},
  pages 157--163.

\bibitem[{Sennrich et~al.(2016)Sennrich, Haddow, and
  Birch}]{sennrich2016improving}
Rico Sennrich, Barry Haddow, and Alexandra Birch. 2016.
\newblock Improving neural machine translation models with monolingual data.
\newblock In \emph{Proceedings of the 54th Annual Meeting of the Association
  for Computational Linguistics (Volume 1: Long Papers)}, pages 86--96.

\bibitem[{Sennrich and Zhang(2019)}]{sennrich2019revisiting}
Rico Sennrich and Biao Zhang. 2019.
\newblock Revisiting low-resource neural machine translation: A case study.
\newblock \emph{arXiv preprint arXiv:1905.11901}.

\bibitem[{Soto et~al.(2020)Soto, Shterionov, Poncelas, and
  Way}]{soto2020selecting}
Xabier Soto, Dimitar Shterionov, Alberto Poncelas, and Andy Way. 2020.
\newblock Selecting backtranslated data from multiple sources for improved
  neural machine translation.
\newblock \emph{arXiv preprint arXiv:2005.00308}.

\bibitem[{Sutskever et~al.(2014)Sutskever, Vinyals, and Le}]{NIPS2014_a14ac55a}
Ilya Sutskever, Oriol Vinyals, and Quoc~V Le. 2014.
\newblock \href
  {https://proceedings.neurips.cc/paper/2014/file/a14ac55a4f27472c5d894ec1c3c743d2-Paper.pdf}
  {Sequence to sequence learning with neural networks}.
\newblock In \emph{Advances in Neural Information Processing Systems},
  volume~27. Curran Associates, Inc.

\bibitem[{Toral(2019)}]{toral2019post}
Antonio Toral. 2019.
\newblock Post-editese: an exacerbated translationese.
\newblock In \emph{Proceedings of Machine Translation Summit XVII Volume 1:
  Research Track}, pages 273--281.

\bibitem[{Vanmassenhove et~al.(2019)Vanmassenhove, Shterionov, and
  Way}]{vanmassenhove2019lost}
Eva Vanmassenhove, Dimitar Shterionov, and Andy Way. 2019.
\newblock Lost in translation: Loss and decay of linguistic richness in machine
  translation.
\newblock In \emph{Proceedings of Machine Translation Summit XVII Volume 1:
  Research Track}, pages 222--232.

\bibitem[{Vaswani et~al.(2017)Vaswani, Shazeer, Parmar, Uszkoreit, Jones,
  Gomez, Kaiser, and Polosukhin}]{vaswani2017attention}
Ashish Vaswani, Noam Shazeer, Niki Parmar, Jakob Uszkoreit, Llion Jones,
  Aidan~N Gomez, {\L}ukasz Kaiser, and Illia Polosukhin. 2017.
\newblock Attention is all you need.
\newblock In \emph{Advances in neural information processing systems}, pages
  5998--6008.

\bibitem[{Yim et~al.(2017)Yim, Joo, Bae, and Kim}]{Yim_2017_CVPR}
Junho Yim, Donggyu Joo, Jihoon Bae, and Junmo Kim. 2017.
\newblock A gift from knowledge distillation: Fast optimization, network
  minimization and transfer learning.
\newblock In \emph{Proceedings of the IEEE Conference on Computer Vision and
  Pattern Recognition (CVPR)}.

\bibitem[{Zhang and Toral(2019)}]{zhang2019effect}
Mike Zhang and Antonio Toral. 2019.
\newblock The effect of translationese in machine translation test sets.
\newblock In \emph{Proceedings of the Fourth Conference on Machine Translation
  (Volume 1: Research Papers)}, pages 73--81.

\end{thebibliography}
\bibliographystyle{acl_natbib}

\newpage
\appendix

\section{Evaluation with CHRF and TER}

\begin{table*}[ht!]

\centering

\begin{tabular}{@{}l@{ ~}l@{ }|@{ }c@{ ~ }c@{ ~ }c@{ }|c@{ ~ }c@{ ~ }c@{ }|c@{ ~ }c@{ ~ }c@{ }|ccc@{}}
\multicolumn{13}{l}{\textbf{CHRF by original language (higher is better)} }                                                                                                                                                                                                                                                                                       \\ 
\hline
& & \multicolumn{9}{c|}{Tr-En model}                                                                                                                                                                                                & \multicolumn{3}{c}{Et-En model}                                           \\
  &                             & \multicolumn{3}{@{ }c@{ }|}{WMT16}                                                & \multicolumn{3}{@{ }c|}{WMT17}                                                & \multicolumn{3}{c|}{WMT18}                                                & \multicolumn{3}{c}{WMT13}                                                 \\
  & Test original language=                              & src                      & tgt                     & all                   & src                      & tgt                      & all                   & src                      & tgt                      & all                   & src                      & tgt                      & all                   \\ 
\hline \hline
 \parbox[t]{2mm}{\multirow{8}{*}{\rotatebox[origin=c]{90}{\small Training Data}}} & \textbf{Big Model}                                                                                           & \multicolumn{1}{l}{}    & \multicolumn{1}{l}{}    & \multicolumn{1}{l|}{} & \multicolumn{1}{l}{} & \multicolumn{1}{l}{}    & \multicolumn{1}{l|}{}    & \multicolumn{1}{l}{} & \multicolumn{1}{l}{} & \multicolumn{1}{l|}{}    & \multicolumn{1}{l}{}    & \multicolumn{1}{l}{} & \multicolumn{1}{l}{}  \\
& ~~\para{} + \btp                             & .478                    & .559                    & .521                  & .471                    & .560                    & .515                  & .455                   & .592                    & .526                  & .550                    & .644                    & .600                  \\
& \textbf{Tiny Model}                     & \multicolumn{1}{l}{}    & \multicolumn{1}{l}{}    & \multicolumn{1}{l|}{} & \multicolumn{1}{l}{}    & \multicolumn{1}{l}{}    & \multicolumn{1}{l|}{} & \multicolumn{1}{l}{}    & \multicolumn{1}{l}{}    & \multicolumn{1}{l|}{} & \multicolumn{1}{l}{}    & \multicolumn{1}{l}{}    & \multicolumn{1}{l}{}  \\
& ~~\para{} + \btp                              & .400                    & .455                    & .429                  & .400                    & .462                    & .430                  & .384                    & .475                    & .431                  & .516                    & .606                    & .564                 \\

& ~~\ft{\para{}+\btp}                              & .452   & .526  & .491                  & .448   & .526  & .487                  & .432   & .552  & .494                  & .531  & .609  & .572                \\
& ~~\ft{\para{}+\mono{}}                              & .469  & .517  & .495                  & .466  & .519   & .492                  & .452  & .546   & .501                  & .551  & .604   & .579                 \\
& ~~\ft{\para{}+\mono{}+\btp}           & .476  & .535  & .508                  & .472  & .538  & .505                  & .454  & .566  & .512                  & .553  & .613  & .584                  \\
& ~~\ft{\para{}+\mono{}}\texttt{+}\btp & .479  & .536 & .509 & .474 & .537  & .505 & .459 & .567 & .515 & .545 & .610 &  .579 \\
\hline
\end{tabular}

\centering

\begin{tabular}{@{}l@{ ~}l@{ }|@{ }c@{ ~ }c@{ ~ }c@{ }|c@{ ~ }c@{ ~ }c@{ }|c@{ ~ }c@{ ~ }c@{ }|ccc@{}}
\multicolumn{13}{l}{\textbf{TER by original language (lower is better)} }                                                                                                                                                                                                                                                                                       \\ 
\hline
& & \multicolumn{9}{c|}{Tr-En model}                                                                                                                                                                                                & \multicolumn{3}{c}{Et-En model}                                           \\
  &                             & \multicolumn{3}{@{ }c@{ }|}{WMT16}                                                & \multicolumn{3}{@{ }c|}{WMT17}                                                & \multicolumn{3}{c|}{WMT18}                                                & \multicolumn{3}{c}{WMT13}                                                 \\
  & Test original language=                              & src                      & tgt                     & all                   & src                      & tgt                      & all                   & src                      & tgt                      & all                   & src                      & tgt                      & all                   \\ 
\hline \hline
 \parbox[t]{2mm}{\multirow{8}{*}{\rotatebox[origin=c]{90}{\small Training Data}}} & \textbf{Big Model}                                                                                           & \multicolumn{1}{l}{}    & \multicolumn{1}{l}{}    & \multicolumn{1}{l|}{} & \multicolumn{1}{l}{} & \multicolumn{1}{l}{}    & \multicolumn{1}{l|}{}    & \multicolumn{1}{l}{} & \multicolumn{1}{l}{} & \multicolumn{1}{l|}{}    & \multicolumn{1}{l}{}    & \multicolumn{1}{l}{} & \multicolumn{1}{l}{}  \\
& ~~\para{} + \btp                             & .708                    & .618                    & .661                  & .708                    & .631                    & .670                  & .762                    & .571                    & .663                  & .578                      & .495                    &  .535               \\
& \textbf{Tiny Model}                     & \multicolumn{1}{l}{}    & \multicolumn{1}{l}{}    & \multicolumn{1}{l|}{} & \multicolumn{1}{l}{}    & \multicolumn{1}{l}{}    & \multicolumn{1}{l|}{} & \multicolumn{1}{l}{}    & \multicolumn{1}{l}{}    & \multicolumn{1}{l|}{} & \multicolumn{1}{l}{}    & \multicolumn{1}{l}{}    & \multicolumn{1}{l}{}  \\
& ~~\para{} + \btp                              & .800                    & .765                    & .781                  & .799                    & .760                    & .780                  & .842                    & .737                    & .788                  & .615                    & .548                    & .581                  \\

& ~~\ft{\para{}+\btp}                              & .722   & .657  & .688                  & .722   & .670  & .696                  & .776   & .630  & .701                 & .599  & .542  & .570                 \\

& ~~\ft{\para{}+\mono{}}                              & .707  & .687  & .696                  & .733  & .685   & .709                  & .773  & .635   & .702                  & .581  & .561   & .570                  \\
& ~~\ft{\para{}+\mono{}+\btp}           &  .698 & .648  & .672           & .701  & .655  & .678                  & .751  & .601  & .673                  & .576 & .536 &  .555                 \\
& ~~\ft{\para{}+\mono{}}\texttt{+}\btp & .698  & .651  & .673 & .704 & .662  & .684 & .752 & .608 & .677 & .589 & .540 & .564 \\
\hline
\end{tabular}

\caption{Experiment results in terms of TER and CHRF, divided based on the test-set's original language. This is the same experiment as Table~\ref{main-res}. Generally, they follow the same pattern as BLEU whereas \texttt{FT(Mono)} improves source-originated test, and  \texttt{FT(BT)} helps target-originated test.}

\end{table*}

To see whether our findings is consistent with other metrics, we re-evaluate our result in Table~\ref{main-res} with CHRF and TER, in which we find a consistent pattern, compared to the result with BLEU.

\end{document}